\documentclass[runningheads]{llncs}

\usepackage{amssymb,amsmath}
\usepackage{blindtext}
\usepackage{graphicx}
\usepackage[misc]{ifsym}
\usepackage{url}
\usepackage{pifont}
\urldef{\mailsa}\path|{venkatesh.n.murthy@siemens-healthineers.com}|   
\newcommand{\keywords}[1]{\par\addvspace\baselineskip
	\noindent\keywordname\enspace\ignorespaces#1}
\graphicspath{ {Img/} }
\usepackage{ctable}

\newcommand{\ie}{{\it i.e.}}

\newcommand{\cmark}{\ding{51}}%
\newcommand{\xmark}{\ding{55}}%

\usepackage{multirow}
\usepackage[normalem]{ulem}

\usepackage{subcaption}
\usepackage{comment}

\usepackage{cite}

\begin{document}
	
	\mainmatter  % start of an individual contribution

	\title{ConTrack: Contextual Transformer for Device Tracking in X-ray}
	\author{Marc Demoustier, Yue Zhang, Venkatesh Narasimha Murthy\textsuperscript{(\Letter)}, Florin C. Ghesu, and Dorin Comaniciu}
	% index{Demoustier, Marc}
	% index{Zhang, Yue}
	% index{Narasimha Murthy, Venkatesh}
	% index{Ghesu, Florin-Cristian}
	% index{Comaniciu, Dorin}
	
	\institute{Digital Technology and Innovation, Siemens Healthineers, Princeton, NJ, USA\\
		\mailsa
	}
	
	\authorrunning{M. Demoustier et al.}
	\titlerunning{Contextual Transformer for Device Tracking in X-ray}

	
	%\toctitle{Lecture Notes in Computer Science}
	%\tocauthor{Authors' Instructions}
	\maketitle

	\begin{abstract}
		    Device tracking is an important prerequisite for guidance during endovascular procedures. Especially during cardiac interventions, detection and tracking of guiding the catheter tip in 2D fluoroscopic images is important for applications such as mapping vessels from angiography (high dose with contrast) to fluoroscopy (low dose without contrast). Tracking the catheter tip poses different challenges: the tip can be occluded by contrast during angiography or interventional devices; and it is always in continuous movement due to the cardiac and respiratory motions. To overcome these challenges, we propose ConTrack, a transformer-based network that uses both spatial and temporal contextual information for accurate device detection and tracking in both X-ray fluoroscopy and angiography. The spatial information comes from the template frames and the segmentation module: the template frames define the surroundings of the device, whereas the segmentation module detects the entire device to bring more context for the tip prediction. Using multiple templates makes the model more robust to the change in appearance of the device when it is occluded by the contrast agent. The flow information computed on the segmented catheter mask between the current and the previous frame helps in further refining the prediction by compensating for the respiratory and cardiac motions. The experiments show that our method achieves 45\% or higher accuracy in detection and tracking when compared to state-of-the-art tracking models.
		    
	\end{abstract}

\section{Introduction}

% 1 Problem
% 2 other works
% 3 Our method
% 4 Contribution
% % Spatio temporal stuffs
% % Transformer for medical tracking
% % Better tracking performance

% Visual object tracking is a fundamental research topic in computer vision and has many uses in modern health care. Object tracking methods have made significant progress in previous years with first Convolution Neural Networks, and then Transformers \cite{vaswani_attention_2017}. Transformers networks solve cases where convolutional networks are bad: long-range dependencies, either in space or time. These tracking models can be leveraged in the medical field for device tracking.

\begin{figure}[t]
	\centering
	\begin{subfigure}{0.32\linewidth}
		\centering
		\includegraphics[scale=0.28]{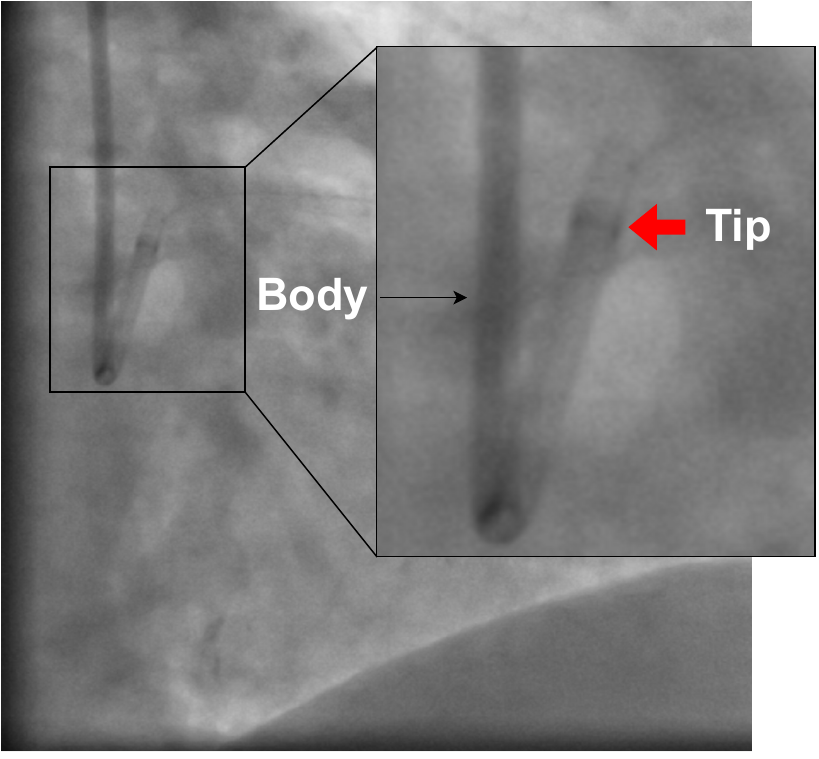}
		\caption{Fluoroscopy} % XA-1500209192472-20111216-1-012-b-00001
		\label{fig:fluoro}
	\end{subfigure}
	\hfill
	\begin{subfigure}{0.32\linewidth}
		\centering
		\includegraphics[scale=0.28]{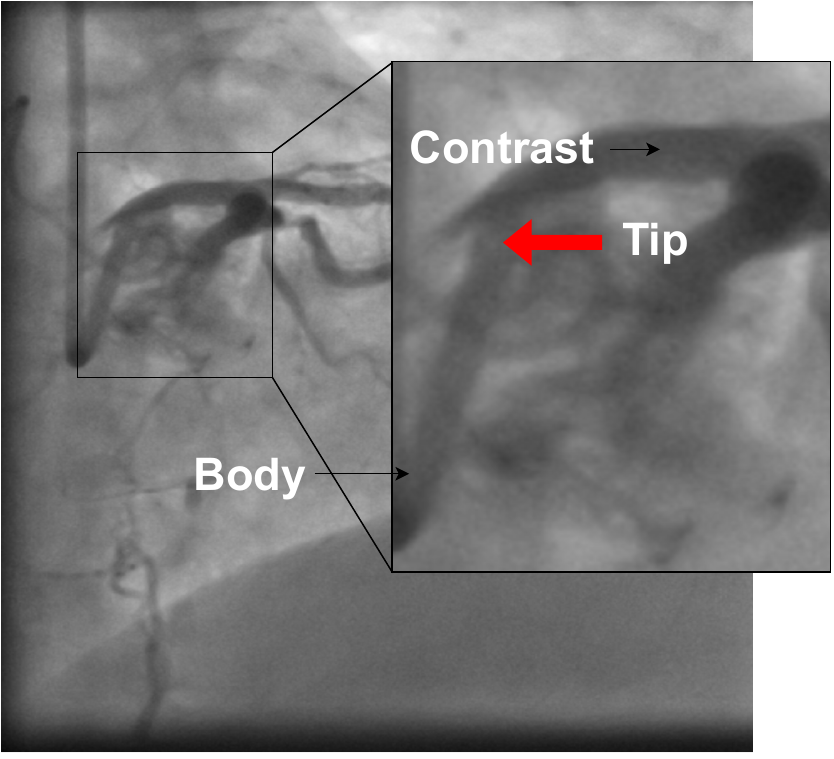}
		\caption{Angiography}
		\label{fig:angio}
	\end{subfigure}
	\hfill
	\begin{subfigure}{0.32\linewidth}
		\centering
		\includegraphics[scale=0.28]{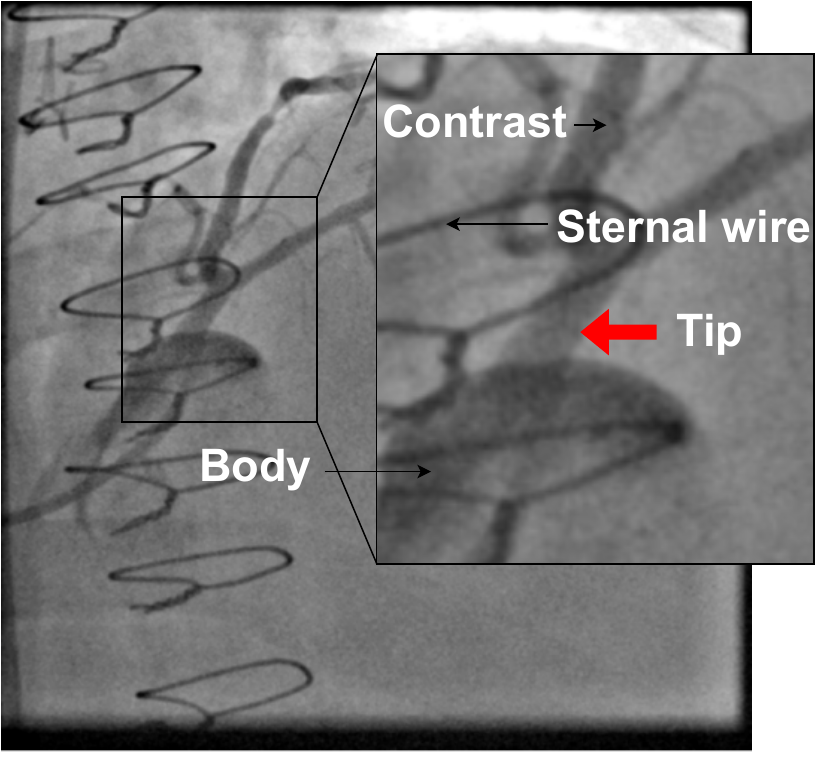}
		\caption{Extra devices} % 20120208-RC-031-01_Angio
		\label{fig:wire}
	\end{subfigure}
	\hfill
	\caption{Example frames from X-ray sequences showing the catheter tip: (a) Fluoroscopy image; (b) Angiographic image with injected contrast medium; (c) Angiographic image with sternum wires. Tracking the tip in angiography is challenging due to occlusion from surrounding vessels and interferring devices.}
	\label{fig:data_comp}
\end{figure}

Tracking of interventional devices plays an important role in aiding surgeons during catheterized interventions such as percutaneous coronary interventions (PCI), cardiac electrophysiology (EP), or trans arterial chemoembolization (TACE). In cardiac image-guided interventions, surgeons can benefit from visual guidance provided by mapping vessel information from angiography (Fig. \ref{fig:angio}) to fluoroscopy (Fig. \ref{fig:fluoro}) \cite{ma_dynamic_2020, wang_image-based_2011} for which the catheter tip is used as an anchor point representing the root of the vessel tree structure. This visual feedback helps in reducing the contrast usage \cite{piayda_dynamic_2018} for visualizing the vascular structures and it can also aid in effective placements of stents or balloons.

Recently, deep learning-based siamese networks have been proposed for medical device tracking \cite{bromley_signature_1993, li_high_2018, lin_cycle_2020}. These networks achieve high frame rate tracking, but are limited by their online adaptability to changes in target’s appearance as they only use spatial information. Cycle Ynet \cite{lin_cycle_2020} uses the cycle consistency of a sequence and relies on a semi-supervised learning approach by doing a forward and a backward tracking. In practice, this method suffers from drifting for long sequences and cannot recover from misdetections because of the single template usage. The closest work related ours is \cite{ma_dynamic_2020}, they use a convolutional neural network (CNN) followed by particle filtering as a post processing step. The drawback of this method is that, it does not compensate for the cardiac and respiratory motions as there is no explicit motion model for capturing temporal information. A similar method adds a graph convolutional neural network for aggregating both spatial information and appearance features \cite{huang_robust_2022} to provide a more accurate tracking but its effectiveness is limited by its vulnerability to appearance changes and occlusion resulting from detection techniques. Optical flow based network architectures \cite{teed_raft_2020} utilize keypoint tracking throughout the entire sequence to estimate the motion of the whole image. However, such approaches are not adapted for tracking a single point, such as a catheter tip.

For general computer vision applications, transformer \cite{vaswani_attention_2017} based-trackers have achieved state-of- the-art performance \cite{yan_learning_2021, yan_towards_2022, cui_mixformer_2022}. Initially proposed for natural language processing (NLP), Transformers learn the dependencies between elements in a sequence, making it intrinsically well suited at capturing global information. Thus, our proposed model consists of a transformer encoder that helps in capturing the underlying relationship between template and search image using self and cross attentions, followed by multiple transformer decoders to accurately track the catheter tip. 

To overcome the limitations of existing works, we propose a generic, end-to-end model for target object tracking with both spatial and temporal context. Multiple template images (containing the target) and a search image (where we would identify the target location, usually the current frame) are input to the system. The system first passes them through a feature encoding network to encode them into the same feature space. Next, the features of template and search are fused together by a fusion network, i.e., a vision transformer. The fusion model builds complete associations between the template feature and search feature and identifies the features of the highest association. The fused features are then used for target (catheter tip) and context prediction (catheter body). While this module learns to perform these two tasks together, spatial context information is offered implicitly to provide guidance to the target detection. In addition to the spatial context, the proposed framework also leverages the temporal context information which is generated using a motion flow network. This temporal information helps in further refining the target location.

Our main contributions are as follows: 
1) Proposed network consists of segmentation branch that provides spatial context for accurate tip prediction; 2) Temporal information is provided by computing the optical flow between adjacent frames that helps in refining the prediction; 3) We incorporate dynamic templates to make the model robust to appearance changes along with the initial template frame that helps in recovery in case of any misdetection; 4) To the best of our knowledge, this is the first transformer-based tracker for real-time device tracking in medical applications; 5) We conduct numerical experiments and demonstrate the effectiveness of the proposed model in comparison to other state-of-the-art tracking models.

%-------------------------------------------------------------------------
\begin{figure}
	\resizebox{\columnwidth}{!}{
		\includegraphics{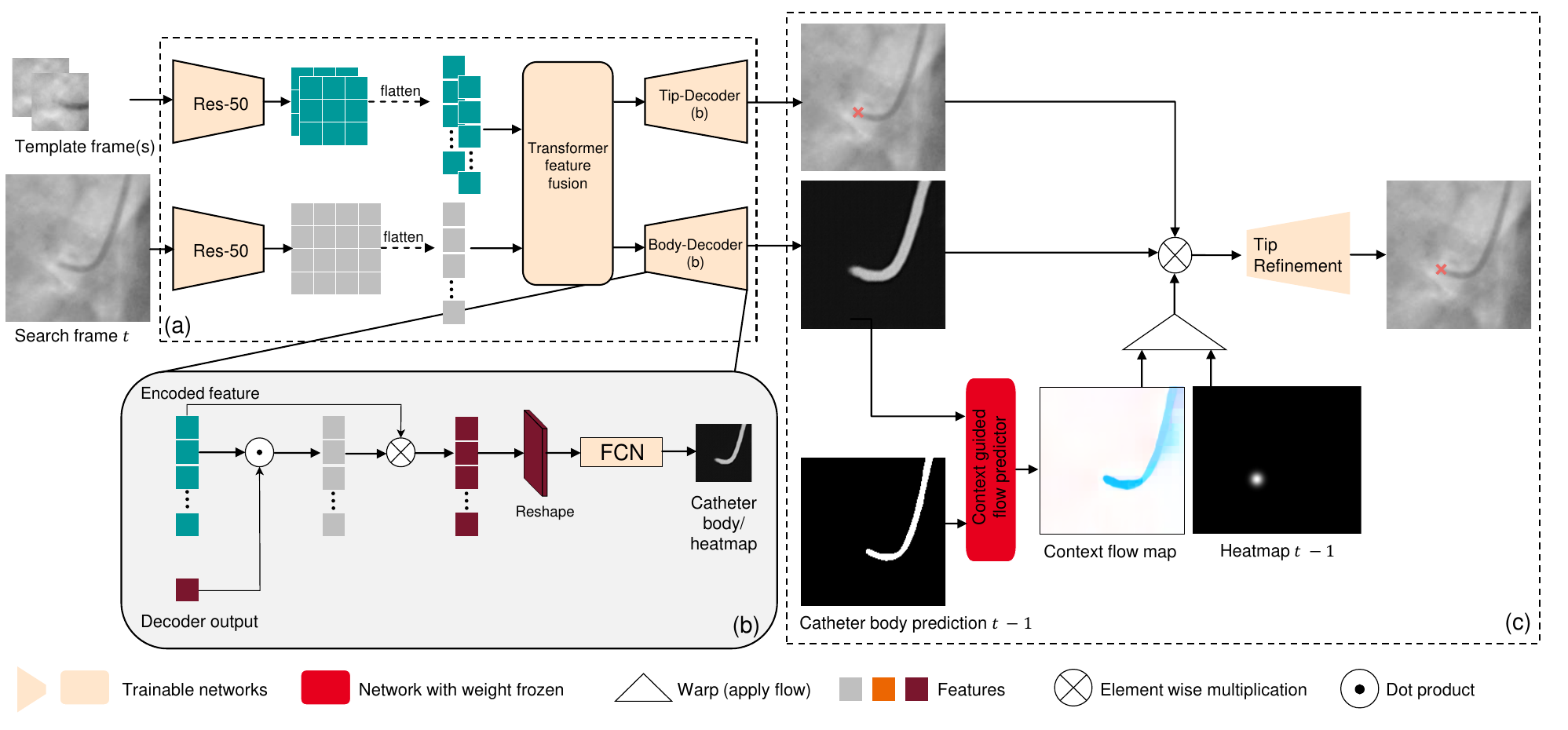}
	}
	\caption{Proposed ConTrack architecture: (a) Transformer feature fusion backbone: ResNet-50 for feature extraction followed by a transformer encoder / decoder; (b) Prediction head for catheter tip (heatmap) and catheter body segmentation (mask segmentation); (c) Flow refinement: use prediction on previous frame to refine the tip detection.}
	\label{fig:contrack}
\end{figure}

\section{Methodology}
Given a sequence of consecutive X-ray images $\{I_t\}_{t=0}^n$ and an initial location of the target catheter tip $x_0 = (u_0, v_0)$, our goal is to track the location of the target $x_t  = (u_t, v_t)$ at any time $t$, $t>0$. The proposed model framework is summarized in Fig. \ref{fig:contrack}. It consists of two stages, target localization stage and motion refinement stage. First, given a selective set of template image patches and the search image, we leverage the CNN-transformer architecture to jointly localize the target and segment the neighboring context, i.e., body of the catheter. Next, we estimate the context motion via optical flow on the catheter body segmentation between neighboring frames and use this to refine the detected target location. We detail these two stages in the following subsections.
%
%First, given a selective set of template image patches and the search image, we obtain their high-level features with share-weighted residual network encoding and leverage transformer encoder to build complete feature point association. Followed by a modified transformer decoder to jointly localize the target and segment the neighboring context, \ie, body of the catheter. Next, we estimate the context motion via optical flow on the catheter body segmentation between neighboring frames and use this to refine the detected target location. Finally, we add confident predictions into the set of templates and use together the context segmentation for target tracking in the next frame.

%
%Throughout the paper, we use subscripts to denote the time stamp$\{z_i\}$, and the search image $I_t$, 

%Following the recent success of transformer-based tracking approaches \cite{yan_learning_2021, mayer2022transforming, cui_mixformer_2022}, we formulate this as a detection problem with temporal constraints. 
%
%Following the recent success of transformer-based tracking approaches[...], we consider the catheter tip tracking problem as an object detection problem. Our proposed model consists of two stages, target localization stage and refinement stage. 

%Given a sequence of . The proposed . We leverage an adaptive set of target templates to 
\subsection{Target localization with multi-template feature fusion}

To identify the target in the search frame, existing approaches build a correlation map between the template and search features. Limited by definition, the template is a single image, either static or from the last frame tracked result. A transformer naturally extends the bipartite relation between template and search images to complete feature associations which allow us to use multiple templates. This improves model robustness against suboptimal template selection which can be caused by target appearance changes or occlusion.

\textbf{Feature fusion with multi-head attention.} In the encoding stage, given a set of template image patches centered around the target $\{T_{ti}\}_{ti\in\mathcal{H}}$ and current frame  $I_s$ as the search image, we aim to determine the target location by fusing information from multiple templates. $\mathcal{H}$ is the set containing historically selected frames for templates. This can be naturally accomplished by multi-head attention (MHA). Specifically, let us denote the ResNet encoder by $\theta$, given the feature map of the search image $\theta(I_s)\in \mathbb{R}^{C\times H_s\times W_s}$, and the feature maps of the templates $\{\theta(T_{ti})\}\}$, we use 1x1 convolutions to project and flatten them into $d-$dimensional vector query, key and value embedding, $q_s$, $k_s$, $v_s$ for the search image features and \{$q_{ti}$\}, \{$k_{ti}$\}, \{$v_{ti}$\} for templates features respectively. The attention is based on the concatenated vectors,
\begin{equation}
	\text{Attention(Q, K, V)} := \text{softmax} (\frac{QK^T}{\sqrt{d}})V,
\end{equation}
where $Q=\text{Concat}(q_s, q_{t1}, q_{t2},..., q_{tn})$, $K=\text{Concat}(k_s, k_{t1}, k_{t2},..., k_{tn})$, $V=\text{Concat}(v_s, v_{t1}, v_{t2},..., v_{tn})$. The definition of MHA then follows \cite{vaswani_attention_2017}.

\textbf{Joint target localization and context segmentation.}
In the decoding stage, we follow \cite{yan_learning_2021} and adjust the transformer decoder to a multi-task setting. As the catheter tip represents a sparse object in the image, solely detecting it suffers from class imbalance issue. To guide the catheter tip tracking with spatial information, we incorporate additional contextual information by simultaneously segmenting the catheter body in the same frame. Specifically, two object queries $(e_1, e_2)$ are employed in the decoder, where $e_1$ defines the position of the catheter tip, and $e_2$ defines the mask of the catheter body. As is illustrated in Fig. \ref{fig:contrack} (b), we first calculate similarity scores between decoder and the encoder output via dot product. We then use element-wise product between the similarity scores and the encoder features to promote regions with high similarity. After reshaping the processed features to $d\times H_s\times W_s$, an encoder-decoder structured 6-layer FCN is attached to process the features to probability maps with the same size as the search image.  A combination of the binary cross-entropy and the dice loss is then used,

\begin{equation}
	\begin{split}
		L =& \lambda^x_{bce} L_{bce}(G(x_i; \mu, \sigma), \hat{x}^s_i) + \lambda^x_{dice} L_{dice}(G(x_i; \mu, \sigma), \hat{x}^s_i) +\\ &\lambda^m_{bce} L_{bce}(m_i, \hat{m}_i) + \lambda^m_{dice} L_{dice}(m_i, \hat{m}_i),
	\end{split}
\end{equation}

where $x_i$, $m_i$ represent the ground truth annotation of the catheter tip and mask, $\hat{x}^s_i$, $\hat{m}^s_i$ are predictions respectively. Here we use sup-script ``$s$'' to denote the predictions from this spatial stage.  $G(x_i; \mu, \sigma) := \exp(-\|x_i-\mu\|^2/\sigma^2)$ is the smoothing function that transfers dot location of $x_i$ to probability map.   $\lambda^*_{bce}, \lambda^*_{dice} \in \mathbb{R}$ are hyperparameters that are empirically optimized. 

%
%A fully convolutional network (FCN) head is applied to this second target query to obtain the segmentation mask. Unlike the previous FCN head, this updated version employs up-convolution to return the segmentation mask to the size of the search frame.
%As for the catheter body segmentation, we use 
%\textbf{Online template selection.}

%\textbf{Training.} The first part of the network remains unchanged. The heatmap prediction head is updated with up-convolution to produce a heatmap of the same size as the search frame. Without gradient computation, the network predicts the position and catheter body on frame t-1, followed by frame t with gradient computation. The optical flow between the two catheter body is calculated using RAFT \cite{teed_raft_2020} with frozen weights and applied to warp the heatmap from frame t-1. The last step is to refine the prediction. Like before, we compute the loss on the heatmap and the mask and on the refined prediction coordinates with the MSE loss.
%The multi-templates and multi-task model feature provides strong spatial guidance for target localization, especially against defective template selection.
\subsection{Localization refinement with context flow}

In interventional procedures, one common challenge for visual tracking comes from occlusion. This can be caused by injected contrast medium (in the angiographic image) or interferring devices such as sternal wires, stent and additional guiding catheters. If the target is occluded in the search image, using only spatial information for localization is inadequate. To address this challenge, we impose a motion prior of the target to further refine the tracked location. As the target is a sparse object, this is done via optical flow estimation of the context.

\textbf{Context flow estimation}. Obtaining ground truth optical flow in real world data is a challenging task and may require additional hardware such as motion sensors. As such, training a model for optical flow estimation directly in the image space is difficult. Instead, we propose to estimate the flow in the segmentation space, i.e., on the predicted heatmaps of the catheter body between neighboring frames. We use the RAFT \cite{teed_raft_2020} model for this task. Specifically, given the predicted segmentation maps $m_{t-1}$ and $m_t$, we first use a 6-block ResNet encoder $g_\theta$ to extract the features $g_\theta(m_{t-1}), g_\theta(m_t) \in R^{H_f\times W_f \times D_f}$. Then we construct the correlation volume pyramid $\{C_i\}_{i=0}^3$, where 
\begin{equation}
	C_i = \text{AvgPool}(corr(g_\theta(m_{t-1}), g_\theta(m_t)), \text{stride}=2^i).
\end{equation}
Here $corr(g_\theta(m_{t-1}), g_\theta(m_t)) \in \mathbb{R}^{H_f\times W_f \times H_f \times W_f} $ stands for correlation evaluation:
\begin{equation}
	corr(g_\theta(m_{t-1}), g_\theta(m_t))_{ijkl} = \sum_{h=1}^{D_f} g_\theta(m_{t-1})_{ijh}  \cdot g_\theta(m_t)_{klh},
\end{equation}
which can be computed via matrix multiplication. Starting with an initial flow $f_0=0$, we follow the same model setup as \cite{teed_raft_2020} to recurrently refine the flow estimates to $f_k = f_{k-1} + \bigtriangleup f$ with a gated recurrent unit (GRU) and a delta flow prediction head of 2 convolutional layers. Given the tracked tip result from the previous frame $\hat{x}_{t-1}$, we can then predict the new tip location at time $t$ by warpping with the context flow $\hat{x}^f_t = f^k (\hat{x}_{t-1})$. Here we use sup-script ``$f$'' to denote the prediction by flow warpping.

We note here that since the segmentations of the catheter body are sparse objects compared to the entire image, computation of the correlation volume and subsequent updates can be restricted to a cropped sub-image which reduces computation cost and flow inference time. As the flow estimation is performed on the segmentation map, one can simply generate synthetic flows and warp them with the existing catheter body annotation to generate data for model training. 

\textbf{Refinement with combined spatial-temporal prediction}. Finally, we generate a score map with combined information from the spatial localization stage and the temporal prediction by context flow,

\begin{equation}
	S_t(u, v) = \begin{cases}
		(\alpha + \hat{m}_t^s(u,v)) (\hat{x}_t^s(u, v) + \hat{x}_t^f(u,v)) & \hat{m}_t^s(u,v) > 0, \\
		\hat{x_t^s}(u, v) + \hat{x}_t^f(u,v) &  \text{otherwise.}
	\end{cases}
\end{equation}

Here $\alpha$ is a positive scalar. It helps the score map to promote coordinates that are activated jointly on both the spatial prediction $\hat{x}_t^s$ and the temporal prediction $\hat{x}_t^f$. Finally, we forward the score map through a refinement module to finalize the prediction. The refinement module consists of a stack of 3 convolutional layers. Similar to the spatial localization stage,  a combination of the binary cross-entropy and the dice loss is used as the final loss.

\section{Experiments and Results}

\textbf{Dataset.} Our study uses an internal dataset of X-ray sequences captured during percutaneous coronary intervention procedures, featuring a field of view displaying the catheter within the patient's heart. The test dataset is divided into two primary categories: fluoroscopic and angiographic sequences. Fluoroscopic sequences are real-time videos of internal movements captured by low-dose X-rays without radiopaque substances, while angiographic sequences display blood vessels in real-time after the introduction of radiopaque substances.

We further separate the test dataset into a third category, ``devices'', presenting a unique challenge for both fluoroscopic and angiographic sequences. In these cases, devices such as wires can obscure the catheter tip and have a similar appearance to the catheter, making tracking more challenging.

The dataset includes frames annotated with the coordinates of the catheter tip and, in some cases, a catheter body mask annotation. For training and validation, we use 2,314 sequences consisting of 198,993 frames, of which 44,957 are annotated. As the model training only requires image pairs, \ie templates and search images, in order to reduce annotation effort, a nonadjacent subset of frames in each sequence is annotated. Their neighboring unannotated frames are also used to provide flow estimation, as is shown in Fig. \ref{fig:contrack}(c). For testing, we use 219 sequences consisting of 17,988 frames, all annotated. The test dataset split is as follows: Fluoro (i.e., fluoroscopy), consisting of 94 sequences, 8,494 frames, from 82 patients; Angio (i.e., angiography), consisting of 101 sequences, 6,904 frames, from 81 patients; and devices, consisting of 24 sequences, 2,593 frames, from 10 patients. All frames undergo a preprocessing pipeline with resampling and padding to size of $512\times512$ with 0.308 mm isotropic pixel spacing.

\textbf{Training.} The template frame is of size $64\times64$. The search frame is of size $160\times160$. With this, the inference speed reaches 12 fps. We train our model for 300 epochs using a learning rate of 0.0001.

\textbf{Comparison study}. We compare the proposed approach with existing arts and summarize the results in Table. \ref{tab:models_comparison}. The proposed approach achieves best performance in all testing dataset. In contrast to our method, SiameseRPN \cite{li_high_2018}, STARK \cite{yan_learning_2021} and MixFormer \cite{cui_mixformer_2022} focus on spatial localization of the target. Temporal information is being incorporated only with the setting of multi-templates thus target motion modeling is limited. While such approaches can achieve good performance with low median errors ($\sim$2mm), the high 5-7mm standard deviations indicate the stability issues, especially in data with devices where occlusions are present. Cycle Ynet \cite{lin_cycle_2020} uses cycle-consistency loss for motion learning directly on the target. As catheter tip is a sparse object, our approach leverages the motion information of the neighboring context which provide more robust guidance for target location refinement.

\begin{figure}[t]
	\centering
	\resizebox{\columnwidth}{!}{
		\includegraphics{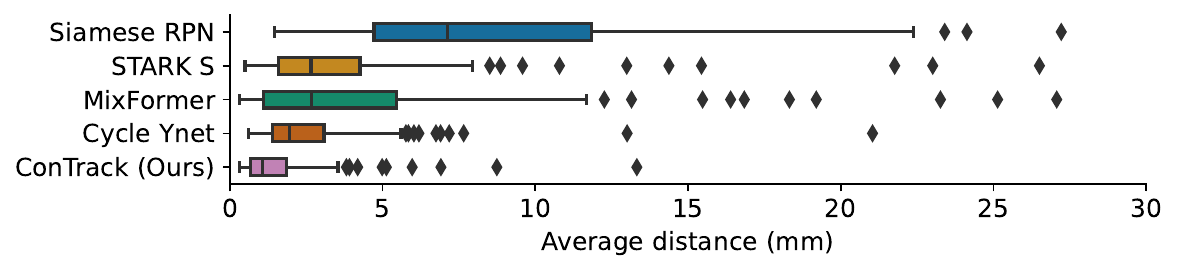}
	}
	\caption{Benchmark study on average distance distribution over all test datasets.}
	\label{fig:percentiles}
\end{figure}

Overall, ConTrack outperforms all other methods, with a median tracking error of less than 1.08mm. Our model is particularly effective at tracking the catheter tip when other devices are in the field of view, where all other methods tend to underperform. Compared to Cycle Ynet on all test datasets, our model is 45\% more accurate, with an average distance of less than 1mm between the prediction and ground truth. Further, we show the accuracy distributions in Fig. \ref{fig:percentiles}. It can be seen that the proposed approach shows superior performance to all other approaches in various percentiles.
\begin{table}
	\centering
	\caption{Comparison study on different testing set. The results are the average distance in mm. Best numbers are marked in bold. Accuracy improvement is statistically significant (p-value $< 0.005$) over the second best in the table.}
	\begin{tabular}{c|cc|cc|cc|ccc|}
		\multirow{2}{*}{Models}             & \multicolumn{2}{c|}{Fluoroscopy}                   & \multicolumn{2}{c|}{Angiography}                   & \multicolumn{2}{c|}{Devices}                       & \multicolumn{3}{c|}{All}                                                                \\
		& \multicolumn{1}{c|}{median}        & mean          & \multicolumn{1}{c|}{median}        & mean          & \multicolumn{1}{c|}{median}        & mean          & \multicolumn{1}{c|}{median}        & \multicolumn{1}{c|}{mean}          & std           \\ \hline
		SiameseRPN \cite{li_high_2018}      & \multicolumn{1}{c|}{6.93}          & 8.19          & \multicolumn{1}{c|}{7.74}          & 9.42          & \multicolumn{1}{c|}{7.89}          & 10.51         & \multicolumn{1}{c|}{7.13}          & \multicolumn{1}{c|}{9.01}          & 6.81          \\ \hline
		STARK \cite{yan_learning_2021}      & \multicolumn{1}{c|}{2.38}          & 3.02         & \multicolumn{1}{c|}{2.82}          & 4.49         & \multicolumn{1}{c|}{4.35}          & 7.01         & \multicolumn{1}{c|}{2.65}          & \multicolumn{1}{c|}{4.14}         & 4.93         \\ \hline
		MixFormer \cite{cui_mixformer_2022} & \multicolumn{1}{c|}{2.02}          & 4.42          & \multicolumn{1}{c|}{2.76}          & 4.86          & \multicolumn{1}{c|}{5.00}          & 9.20          & \multicolumn{1}{c|}{2.68}          & \multicolumn{1}{c|}{5.15}          & 7.10          \\ \hline
		Cycle Ynet \cite{lin_cycle_2020}    & \multicolumn{1}{c|}{2.05}          & 2.92          & \multicolumn{1}{c|}{1.69}          & 2.09          & \multicolumn{1}{c|}{4.39}          & 4.23          & \multicolumn{1}{c|}{1.96}          & \multicolumn{1}{c|}{2.68}          & 2.40          \\ \hline
		\textbf{ConTrack (Ours)}            & \multicolumn{1}{c|}{\textbf{0.73}} & \textbf{1.04} & \multicolumn{1}{c|}{\textbf{1.27}} & \textbf{1.91} & \multicolumn{1}{c|}{\textbf{1.61}} & \textbf{2.73} & \multicolumn{1}{c|}{\textbf{1.08}} & \multicolumn{1}{c|}{\textbf{1.63}} & \textbf{1.70} \\ \hline
	\end{tabular}
	\label{tab:models_comparison}
\end{table}

\textbf{Ablation study.} We conduct an ablation study to investigate the effectiveness of different model components. Results are summarized in Table \ref{tab:ablation}. Our ablation study revealed three key findings: 1) The addition of the mask segmentation branch improved tracking performance on Fluoro, where the device appearance remains consistent and there is no occlusion. However, when there are distractors, the results are less accurate; 2) The inclusion of a mask segmentation enabled the estimation of motion. The resulting flow helped to stabilize tracking in the presence of distractors; and 3) Multiple templates were employed to better handle changes in appearance. The combined model showed the best performance in dataset of angiography and data with devices, while yielding similar results in dataset of fluoroscopy.
\begin{table}[t]
	\centering
	\caption{Ablation study on model components. \xmark \ denotes the component is removed, while \cmark \ represents the component is used. Performance is evaluated on the same test cases as before and the results are the average distance in mm.}
	\resizebox*{\textwidth}{!}{%
		\begin{tabular}{c|c|c|cc|cc|cc|ccc|}
			\multirow{2}{*}{Multitask} & \multirow{2}{*}{Flow}     & \multirow{2}{*}{\begin{tabular}[c]{@{}c@{}}Multi-\\templates\end{tabular}} & \multicolumn{2}{c|}{Fluoro}                        & \multicolumn{2}{c|}{Angio}                         & \multicolumn{2}{c|}{Devices}                       & \multicolumn{3}{c|}{All}                           \\
			&                           &                                                                             & \multicolumn{1}{c|}{median}                             & mean          & \multicolumn{1}{c|}{median}                             & mean          & \multicolumn{1}{c|}{median}                             & mean          & \multicolumn{1}{c|}{median}                             & \multicolumn{1}{c|}{mean}          & std          \\ \hline
			\xmark                          & \xmark                         & \xmark                                                                           & \multicolumn{1}{c|}{0.81}        & 1.40          & \multicolumn{1}{c|}{1.29}          & 1.94 & \multicolumn{1}{c|}{2.99}          & 6.20         & \multicolumn{1}{c|}{1.13}          & \multicolumn{1}{c|}{2.17}        & 3.75          \\ \hline
			\cmark  & \xmark                         & \xmark                                                                           & \multicolumn{1}{c|}{0.67}        & 1.03          & \multicolumn{1}{c|}{1.53}          & 2.15          & \multicolumn{1}{c|}{3.97}          & 10.49         & \multicolumn{1}{c|}{1.11}          & \multicolumn{1}{c|}{2.58}         & 6.10          \\ \hline
			\cmark  & \cmark & \xmark                                                                           & \multicolumn{1}{c|}{\textbf{0.65}} & \textbf{0.96}          & \multicolumn{1}{c|}{1.49}          & 1.95          & \multicolumn{1}{c|}{1.93}          & 4.52         & \multicolumn{1}{c|}{\textbf{0.99}} & \multicolumn{1}{c|}{1.81}          & 2.30          \\ \hline
			\cmark  & \cmark & \cmark                                                   & \multicolumn{1}{c|}{0.73}        & 1.05 & \multicolumn{1}{c|}{\textbf{1.27}} & \textbf{1.91}          & \multicolumn{1}{c|}{\textbf{1.61}} & \textbf{2.73} & \multicolumn{1}{c|}{1.08}          & \multicolumn{1}{c|}{\textbf{1.63}} & \textbf{1.70} \\ \hline
		\end{tabular}
	}
	\label{tab:ablation}
\end{table}

Despite our framework's incorporation of various temporal and spatial contexts, catheter tracking remains a challenging task, particularly in cases where other devices or contrast agents obscure the catheter tip and create visual similarities with the catheter itself. Nonetheless, our results demonstrate the promise of ConTrack as a valuable tool for enhancing catheter tracking accuracy.

\section{Conclusion}

Device tracking is an important task in interventional procedures. In this paper, we propose a generic model framework, ConTrack, that leverages both spatial and temporal information of the surrounding context for accurate target localization and tracking in X-ray.  Through extensive experimentation on large datasets, our approach demonstrated superior tracking performance, outperforming other state-of-the-art tracking models, especially in challenging scenarios where occlusions and distractors are present. Current approach has its limitations. Motion estimation is learned from neighboring two frames and thus target historical trajectory information is missing. Further, transformer-based model training require large amount of annotated data, which is challenging to collect in interventional applications. Finally, throughout the paper we follow established setups and focus on the development on the tracking model with manual initialization. In general, long-term visual tracking with automatic (re-)initialization is a challenging problem and require a system of approaches. A safe and automatic system of device and anatomy tracking is of great clinical relevance and will be an important future work for us.

%the ConTrack model, a transformer-based framework that leverages both spatial (segmentation branch) and temporal (optical flow with multiple templates) contextual information for accurate catheter tip detection and tracking in both X-ray fluoroscopy and angiography. (45\% more accurate than Cycle Ynet \cite{lin_cycle_2020}). These capabilities make ConTrack a powerful tool for real-world medical imaging applications.

\subsubsection{Disclaimer}
The concepts and information presented in this paper/presentation are based on research results that are not commercially available. Future commercial availability cannot be guaranteed.

\bibliographystyle{splncs04}
\bibliography{ref}

\end{document}